\documentclass[runningheads]{llncs}
\usepackage{graphicx}
\usepackage{amsmath,amssymb}
\usepackage{color}

\usepackage{xspace}
\usepackage{booktabs}
\usepackage{wrapfig}
\usepackage[normalem]{ulem}
\makeatletter
\DeclareRobustCommand\onedot{\futurelet\@let@token\@onedot}

\def\@onedot{\ifx\@let@token.\else.\null\fi\xspace}

\usepackage[font=small,labelfont=bf]{caption}
\usepackage{pdflscape}
\usepackage{afterpage}
\usepackage{xcolor,colortbl}

\def\eg{\emph{e.g}\onedot} 
\def\ie{\emph{i.e}\onedot}

\def\etal{\emph{et al}\onedot}
\makeatother
\usepackage{tikz}
\newcommand*{\x}{{\times}}

\newcommand{\flexconvlayer}{\protect\includegraphics{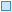} }
\newcommand{\neighbop}{\protect\includegraphics{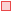} }
\newcommand{\positionop}{\protect\includegraphics{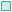} }
\newcommand{\flexdeconvlayer}{\protect\includegraphics{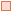} }
\newcommand{\flexmaxlayer}{\protect\includegraphics{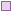} }

\begin{document}
\pagestyle{headings}
\mainmatter

\def\ACCV18SubNumber{453}

\title{Flex-Convolution \newline \normalsize Million-Scale Point-Cloud Learning Beyond Grid-Worlds}

\titlerunning{Flex-Convolution}

\authorrunning{Fabian Groh, Patrick Wieschollek, Hendrik P.A. %
Lensch}

\author{Fabian Groh\inst{1}\orcidID{0000-0001-8717-7535} \and
Patrick Wieschollek\inst{1,2}\orcidID{0000-0002-4961-9280} \and
Hendrik P.A.
Lensch\inst{2}\orcidID{0000-0003-3616-8668}}

\institute{University of T{\"u}bingen, Germany \and
Max Planck Institute for Intelligent Systems, T{\"u}bingen, Germany
}

\maketitle

\begin{abstract}

Traditional convolution layers are specifically designed to exploit the natural data representation of images -- a fixed and regular grid.
However, unstructured data like 3D point clouds containing irregular neighborhoods constantly breaks the grid-based data assumption.
Therefore applying best-practices and design choices from 2D-image learning methods towards processing point clouds are not readily possible.
In this work, we introduce a natural generalization \textit{flex-convolution} of the conventional convolution layer along with an efficient GPU implementation.
We demonstrate competitive performance on rather small benchmark sets using fewer parameters and lower memory consumption and obtain significant improvements on a million-scale real-world dataset.
Ours is the first which allows to efficiently process 7 million points \textit{concurrently}.

 \end{abstract}

\section{Introduction}
\label{sec:introduction}
Deep Convolutional Neural Networks (CNNs) shine on tasks where the underlying data representations are based on a regular grid structure, e.g., pixel representations of RGB images or transformed audio signals using Mel-spectrograms \cite{audioprocessing}.
For these tasks, research has led to several improved neural network architectures ranging from VGG~\cite{vggnet} to ResNet~\cite{resnet}.
These architectures have established state-of-the-art results on a broad range of classical computer vision tasks~\cite{iccv2017/Wieschollek} and effortlessly process entire HD images ($\sim$2 million pixels)
 within a single pass.
This success is fueled by recent improvements in hardware and software stacks (\eg. TensorFlow),
which provide highly efficient implementations of layer primitives \cite{winograd} in specialized libraries~\cite{cudnn} exploiting the grid-structure of the data.
It seems appealing to use grid-based structures (\eg voxels) to process higher-dimensional data relying on these kinds of layer implementations.
However, grid-based approaches are often unsuited for processing irregular point clouds and unstructured data.
The grid resolution on equally spaced grids poses a trade-off between discretization artifacts and memory consumption.
Increasing the granularity of the cells is paid by higher memory requirements that even grows exponentially due to the curse of dimensionality.

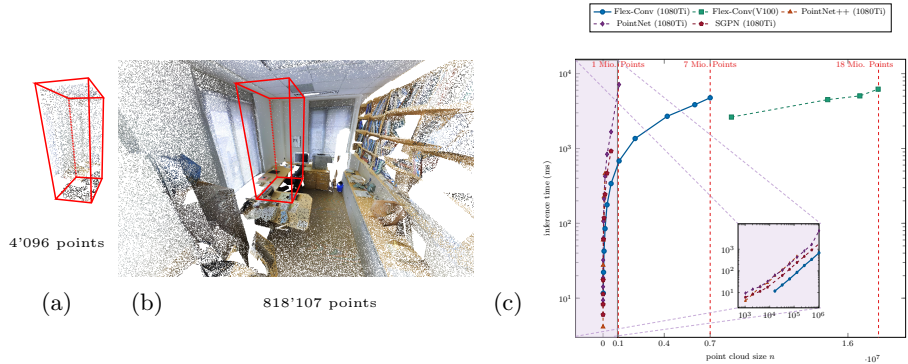
\begin{figure}[tb]
  \centering
  \begin{tikzpicture}
    \node {\includegraphics[width=.5\textwidth]{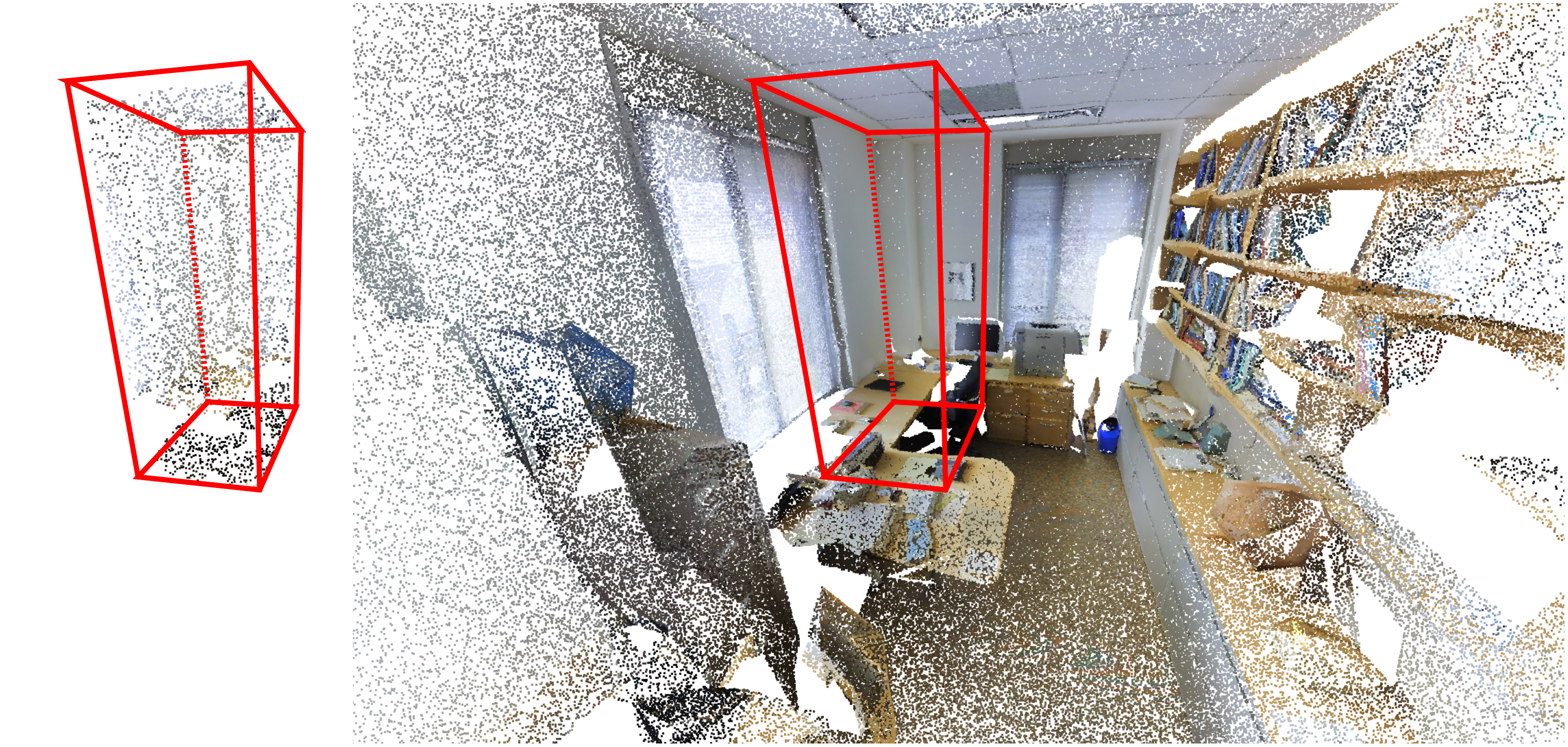}};
    \node at (-2.50,-1) {\tiny 4'096 points};
    \node at (1,-1.8) {\tiny 818'107 points};
    \node at (-2.50,-1.8) {(a)};
    \node at (-1.30,-1.8) {(b)};
    \node at (3.5,-1.8) {(c)};
    \node at (0, -2.5) {};
  \end{tikzpicture}
  \includegraphics[width=.4\textwidth]{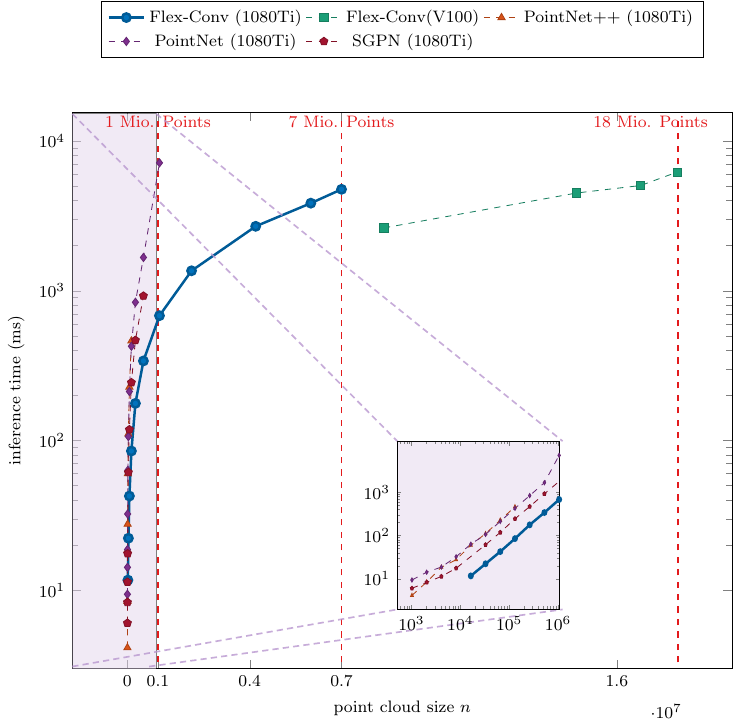}
  \caption{Processing full-resolution point clouds is an important ingredient for successful semantic segmentation.
  Previous methods \cite{pointnet,pointnet2,wang2018sgpn} subsample small blocks (a), while ours (b) processes the entire room and can (c) handle inputs up to 7 Million points in a single forward-pass with the \textit{same} accuracy.
  % \vspace{-1cm}
  Previous methods \textit{could} handle at most 1 Million points -- but training is not feasible on today's hardware.
  }
  \label{fig:teaser}
  \vspace{-0.7cm}
\end{figure}

While training neural networks on 3D voxel grids is possible~\cite{voxnet}, even with hierarchical octrees~\cite{octnet} the maximum resolution is limited to $256^3$ voxels --- large data sets are currently out-of-scope.
Another issue is the discretization and resampling of continuous data into a fixed grid.
For example, depth sensors produce an arbitrarily oriented depth map with different resolution in $x,y$ and $z$.
In Structure-from-Motion, the information of images with arbitrary perspective, orientation and distance to the scene --- and therefore resolution --- need to be merged into a single 3D point cloud.
This potentially breaks the grid-based structure assumption completely, such that processing such data in full resolution with conventional approaches is infeasible by design.
These problems become even more apparent when extending current data-driven approaches to handle higher-dimensional data.
A solution is to learn from unstructured data directly.
Recently, multiple attempts from the PointNet family \cite{pointnet,pointnet2,wang2018sgpn} amongst others \cite{kdnetwork,su18splatnet,hermosilla2018monte} proposed to handle \textit{irregular} point clouds directly in a deep neural network.
In contrast to the widely successful general purpose 2D network architectures, these methods propose very particular network architectures with an optimized design for very specific tasks.
Also, these solutions only work on rather small point clouds, still lacking support for processing million-scale point cloud data.
Methods from the PointNet family subsample their inputs to 4096 points per $1m^2$ as depicted in Figure~\ref{fig:teaser}.
Such a low resolution enables single object classification, where the primary information is in the global shape characteristics ~\cite{modelnet40}.
Dense, complex 3D scenes, however, typically consist of millions of points \cite{3d2dsdata,DBLP:conf/dagm/GrohRL17}.
Extending previous learning-based approaches to \textit{effectively} process larger point clouds has been infeasible (Figure~\ref{fig:teaser} (c)).

Inspired by commonly used CNNs architectures, we hypothesize that a simple convolution operation with a small amount of learnable parameters is advantageous when employing them in deeper network architectures --- against recent trends of proposing complex layers for 3D point cloud processing.

To summarize our main contributions: (1) We introduce a novel convolution layer for arbitrary metric spaces, which represents a natural generalization of traditional grid-based convolution layers along (2) with a highly-tuned GPU-based implementation, providing significant speed-ups.
(3) Our empirical evaluation demonstrates substantial improvements on large-scale point cloud segmentation \cite{3d2dsdata} \textit{without} any post-processing steps, and competitive results on small benchmark sets using fewer parameters and less memory.

 \section{Related Work}
\label{sec:related_work}

Recent literature dealing with learning from 3D point cloud data can be organized into three categories based on their way of dealing with the input data.
\par\textbf{Voxel-based} methods~\cite{voxnet,qi2016volumetric,Zhirong15CVPR,octnet} discretize the point cloud into a voxel-grid enabling the application of classical convolution layers afterwards.
However, this either loses spatial information during the discretization process or requires substantial computational resources for the 3D convolutions to avoid discretization artifacts.
These approaches are affected by the curse of dimensionality and will be infeasible for higher-dimensional spaces.
Interestingly, ensemble methods
\cite{SuMKL15,PANORAMA}
based on classical CNNs still achieve state-of-the-art results on common benchmark sets like ModelNet40 \cite{modelnet40} by rendering the 3D data from several viewing directions as image inputs.
As the rendered views omit some information (\ie occlusions) Cao~\etal~\cite{sphericalprojection} propose to use a spherical projection.
\par
\textbf{Graph-based} methods are geared to process social networks or knowledge graphs, particular instances of unstructured data where each node locations is solely defined by its relation to neighboring nodes in the absence of absolute position information.
Recent research \cite{kipf2017semi} proposes to utilize a sparse convolution for graph structures based on the adjacency matrix.
This effectively masks the output of intermediate values in the classical convolution layers and mimics a diffusion process of information when applying several of these layers.
\par\textbf{Euclidean Space-based} methods deal directly with point cloud data featuring absolute position information but \textit{without} explicit pair-wise relations.
PointNet \cite{pointnet} is one of the first approaches yielding competitive results on ModelNet40.
It projects each point \textit{independently} into some learned features space, which then is transformed by a spatial transformer module \cite{spatialtransformer} -- a rather costly operation for higher feature dimensions.
While the final aggregation of information is done effectively using a max-pooling operation, keeping all high dimensional features in memory beforehand is indispensable and becomes infeasible for larger point clouds by hardware restrictions.
The lack of granularity during features aggregation from local areas is addressed by the extension {PointNet++}~\cite{pointnet2} using ``mini''-PointNets for each point neighborhood across different resolutions and later by \cite{wang2018sgpn}.
An alternative way of introducing a structure in point clouds relies on kD-trees \cite{kdnetwork}, which allows to share convolution layers depending on the kD-tree splitting orientation.
Such a structure is affected by the curse of dimensionality can only fuse point pairs in each hierarchy level.
Further, defining splatting and slicing operations \cite{su18splatnet} has shown promising results on segmenting a facade datasets.
{Dynamic Edge-Condition Filters}~\cite{dyncedgefilters} learn parameters in the fashion of Dynamic Filter-Networks \cite{dynamicfilternetwork} for each single point neighborhood.
Note, predicting a neighborhood-dependent filter can become quickly expensive for reasonably large input data.
It is also noted by the authors, that tricks like BatchNorm are required during training.

Our approach belongs to the third category proposing a natural extension of convolution layers (see next section) for unstructured data which can be considered as a scalable special case of \cite{dynamicfilternetwork} but allows to evaluate point clouds and features more efficiently ``in one go'' -- without the need of additional tricks.

 \section{Method}
\label{sec:method}

The basic operation in convolutional neural networks is a discrete 2D convolution, where the image signal\footnote{$c\in C$ represents the RGB, where we abuse notation and write $C$ for $\{0,1,\ldots, C-1\}\subset \mathbb{N}$ as well.}
$I\in\mathbb{R}^{H\x W\x C}$ is convolved with a filter-kernel $w$.
In deep learning a common choice of the filter size is $3\x 3\x C$ such that this mapping can be described as
\begin{align}
  (w \circledast f) [\ell] = \sum_{c\in C}\sum_{\tau\in \{-1,0,1\}^2} w_{c'}(c, \tau) f(c, \ell - \tau),
  \label{eq:conv_discrete}
\end{align}
where $\tau\in \{-1,0,1\}^2$ describes the 8-neighborhood of $\ell$ in regular 2D grids.
One usually omits the location information $\ell$ as it is given implicitly by arranging the feature values on a grid in a canonical way.
Still, each pixel information is a \textit{pair} of a feature/pixel value $f(c, \ell)$ and its location $\ell$.

In this paper, we extend the convolution operation $\circledast$ to support irregular data with real-valued locations. In this case, the kernel $w$ needs to support arbitrary relative positions $\ell_i-\tau_i$, which can be potentially unbounded.
Before discussing such potential versions of $w$, we shortly recap the grid-based convolution layer in more detail to derive desired properties of a more generic convolution operation.

\subsection{Convolution Layer}

For a discrete $3\x 3\x C$ convolution layer such a filter mapping\footnote{$1_M$ is the indicator function being 1 iff $M\neq \emptyset$.}
\begin{align}
  w_{c'}\colon C\x\{-1,0,1\}^2 \to \mathbb{R},\quad (c, \tau) \mapsto w_{c'}(c, \tau) =\sum_{\tau'\in \{-1,0,1\}^2} 1_{\{\tau= \tau'\}} w_{c, c', \tau'}
\end{align}
is based on a lookup table with $9$ entries for each $(c, c')$ pair. These values $w_{c, c', \tau'}$ of the box-function $w_{c'}$ can be optimized for a specific task, \eg using back-propagation when training CNNs.
Typically, a single convolution layer has a filter bank of multiple filters.
While these box functions are spatially invariant in $\ell$, they have a bounded domain and are neither differentiable nor continuous wrt. $\tau$ by definition.
Specifically, the 8-neighborhood in a 2D grid always has exactly the same underlying spatial layout.
Hence, an implementation can exploit the implicitly given locations. The same is also true for other filter sizes $k_h \x k_w\x C$.

Processing irregular data requires a function $w_{c'}$, which can handle an \textit{unbounded} domain of arbitrary --- potentially real-valued --- relations between $\tau$ and $\ell$, besides retaining the ability to share parameters across different neighborhoods.
To find potential candidates and identify the required properties, we consider a point cloud as a more generic data representation
\begin{align}
  P=\left\{(\ell^{(i)}, f^{(i)})\in L\x F\;|\; i=0,1,\ldots,n-1\right\}.
\end{align}
Besides its value $f^{(i)}$, each point cloud element now carries an \textit{explicitly} given location information $\ell^{(i)}$.
In arbitrary metric spaces, \eg Euclidean space $(\mathbb{R}^d, \Vert{\cdot}\Vert)$, $\ell^{(i)}$ can be real-valued without matching a discrete grid vertex.
Indeed, one way to deal with this data structure is to \textit{voxelize} a given location $\ell\in \mathbb{R}^d$ by mapping it to a specific grid vertex, \eg $L' \subset \alpha\mathbb{N}^d, \alpha \in \mathbb{R}$. When $L'$ resembles a grid structure, classical convolution layers can be used after such a discretization step.
As already mentioned, choosing an appropriate $\alpha$ causes a trade-off between rather small cells for finer granularity in $L'$ and consequently higher memory consumption.

Instead, we propose to define the notion of a convolution operation for a set of points in a local area.
For any given point at location $\ell$ such a set is usually created by computing the $k$ nearest neighbor points with locations $\mathcal{N}_k(\ell)=\{\ell_0', \ell_1',\ldots, \ell_{k-1}'\}$ for a point at $\ell$, e.g.\ using a kD-tree.
Thus, a generalization of Eq.~\eqref{eq:conv_discrete} can be written as
\begin{align}
  f'(c', \ell^{(i)}) = \sum_{c\in C}\sum_{\ell'\in\mathcal{N}_k(\ell^{(i)})} \tilde{w}(c, \ell^{(i)}, \ell') \cdot f(c, \ell').
  \label{eq:generic_conv}
\end{align}
Note, for point clouds describing an image
Eq. \eqref{eq:generic_conv} is equivalent\footnote{By setting $\mathcal{N}_9(\ell)=\{\ell - \tau|\tau\in \{-1,0,1\}^d\}$ and $\tilde{w}_{c'}(c, \ell^{(i)}, \ell') = w_{c'}(c, \ell^{(i)} - \ell')$.}
to Eq. \eqref{eq:conv_discrete}.
But for the more general case we require that
\begin{align}
  \tilde{w}_{c'}\colon C\x\mathbb{R}^d\x\mathbb{R}^d \to \mathbb{R},\quad (c, \ell, \ell') \mapsto \tilde{w}(c, \ell, \ell')
\end{align}
is an \textit{everywhere} well-defined function instead of a ``simple'' look-up table.
This ensures, we can use $\tilde{w}$ in neighborhoods of arbitrary sizes.
However, a side-effect of giving up the grid-assumption is that $\tilde{w}$ needs to be differentiable in both $\ell, \ell'$ to perform back-propagation during training.

While previous work \cite{pointnet2,dyncedgefilters} exert small neural networks for $\tilde{w}$ as a workaround inheriting all previously described issues, we rely on the given standard scalar product as the natural choice of $\tilde{w}$ in the Euclidean space with learnable parameters $\theta_c\in \mathbb{R}^d, \theta_{b_c}\in \mathbb{R}$:
\begin{align}
  \tilde{w}(c, \ell, \ell' | \, \theta_c, \theta_{b_c}) &= \left\langle \theta_c, \ell - \ell'\right\rangle + \theta_{b_c} \, .
  \label{eq:ours}
\end{align}

\begin{figure}[h]
  \centering
  \includegraphics[width=\textwidth]{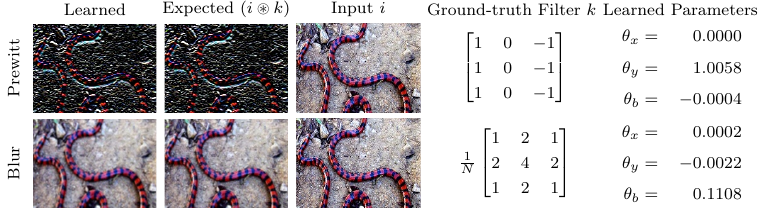}
  \caption{Results on a toy dataset for illustration purposes. The special-case $w(x, y) = \theta_x(x-x_0) +\theta_y(y-y_0) + \theta_{b_c}$ of Eq.~\eqref{eq:ours} is trained to re-produce the results of basic image operations like Prewitt or Blur. }
  \label{fig:toydata}
\end{figure}

This formulation can be considered as a linear approximation of the lookup table, with the advantage of being defined everywhere.
In a geometric interpretation $\tilde{w}$ is a learnable linear transformation (scaled and rotated) of a high-dimensional Prewitt operation.
It can represent several image operations; Two are depicted in Figure~\ref{fig:toydata}.
\clearpage

Hence, the mapping $\tilde{w}$ from Eq.~\eqref{eq:ours} exists in all metric spaces, is \textit{everywhere} well-defined in $c, \ell, \ell'$, and continuously differentiable wrt. to \textit{all} arguments, such that gradients can be propagated back even through the locations $\ell, \ell'$.
Further, our rather simplistic formulation results in a significant reduction of the required trainable parameters and retains translation invariance. %
One observed consequence is a more stable training even \textit{without} tricks like using BatchNorm as in \cite{dyncedgefilters}.
This operation is parallel and can be implemented using CUDA to benefit from the sparse access patterns of local neighborhoods.
In combination with a minimal memory footprint, this formulation is the first being able to process millions of irregular points simultaneously -- a crucial requirement when applying this method in large-scale real-world settings.
% To summarize, the choice of $\tilde{w}$ using the standard scalar product has several advantages:
% \begin{enumerate}
% %   \item The mapping $\tilde{w}$ from Eq.~\eqref{eq:ours} exists in all metric spaces and is \textit{everywhere} well-defined in $c, \ell, \ell'$.
% % Hence, it can cover different distance ranges of neighborhood entries without the need for discretization or value clipping.
% %   \item The mapping $\tilde{w}$ is continuously differentiable wrt. to \textit{all} arguments, such that gradients can be propagated back even through the locations $\ell, \ell'$.
%   \item While $\tilde{w}$ retains translation invariance, compared to previous methods, our rather simplistic formulation results in a significant reduction of the required trainable parameters $\theta_c, \theta_{b_c}$ in $\tilde{w}$ --- besides a minimal memory footprint for storing parameters.
%   One observed consequence is a more stable training even \textit{without} tricks like using BatchNorm as in \cite{dyncedgefilters}.
%   \item This operation is parallel and can be implemented using CUDA to benefit from the sparse access patterns of local neighborhoods.
%   In combination with a minimal memory footprint, this formulation is the first being able to process millions of irregular points simultaneously -- a crucial requirement when applying this method in large-scale real-world settings.
% \end{enumerate}
We experimented with slightly more complex versions of flex-conv, \eg using multiple sets of parameters for one filter dependent on local structure.
However, they did not lead to better results and induced unstable training.

\subsection{Extending Sub-Sampling to Irregular Data}
\label{sec:sub-Sampling}
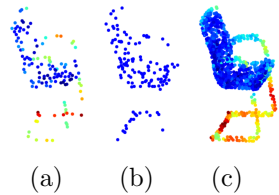
\begin{wrapfigure}{r}{0.41\textwidth}
  \centering
  \begin{tikzpicture}
    \node{\includegraphics[width=.3\textwidth]{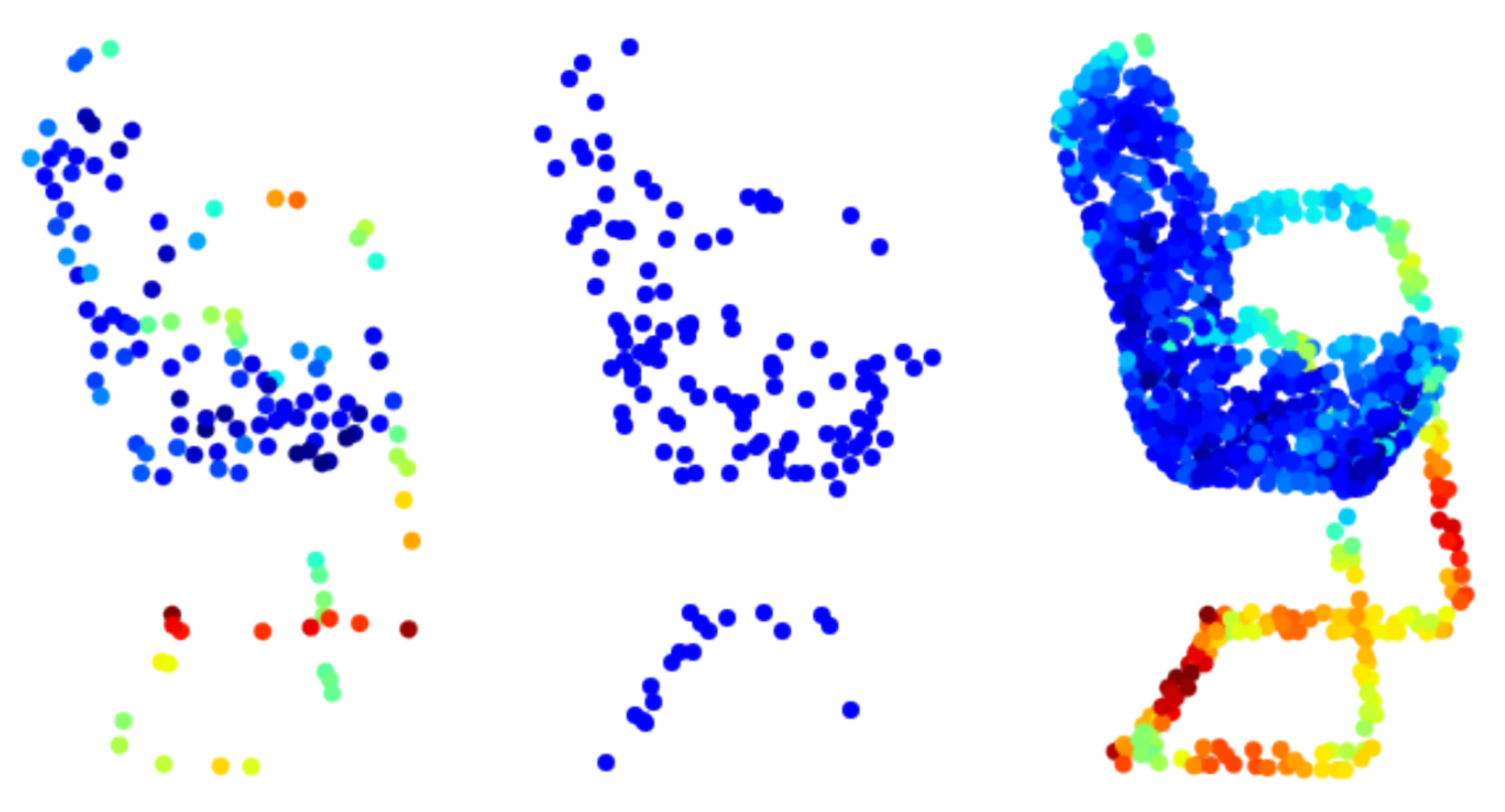}};
    \node at (-1.30,-1.4) {(a)};
    \node at (-1.30+1*1.2,-1.4) {(b)};
    \node at (-1.30+2*1.2,-1.4) {(c)};
  \end{tikzpicture}
  \caption{IDISS (a) against random sub-sampling (b) for an object (c) with color-coded density. }
  \label{fig:subsampling}
  % \vspace{-0.8cm}
\end{wrapfigure}
While straightforward in grid-based methods, a proper and scalable sub-sampling operation in unstructured data is not canonically defined.
On grids, down-sampling an input by a factor 4 is usually done by just taking every second cell in each dimension and aggregating information from a small surrounding region. There is always an implicitly well-defined connection between a point and its representative at a coarser resolution.

For sparse structures this property no longer holds.
Points being neighbors in one resolution, potentially are not in each other's neighborhood at a finer resolution.
Hence, it is even possible that some points will have no representative within the next coarser level.
To avoid this issue, Simonovsky \etal~\cite{dyncedgefilters} uses the VoxelGrid algorithm which inherits all voxel-based drawbacks described in the previous sections.
Qi \etal~\cite{pointnet2} utilizes Farthest point sampling (FPS).
While this produces sub-samplings avoiding the missing representative issue, it pays the price of having the complexity of $\mathcal{O}(n^2)$ for \textit{each} down-sampling layer.
This represents a serious computation limitation.
Instead, we propose to utilize inverse density importance sub-sampling (IDISS). In our approach, the inverse density $\phi$ is simply approximated by adding up all distances from one point in $\ell$ to its $k$-neighbors by $\phi (\ell) = \sum_{\ell' \in \mathcal{N}_k(\ell)} \left\Vert \ell - \ell' \right\Vert$.

Sampling the point cloud proportional to this distribution has a computational complexity of $\mathcal{O}(n)$, and thereby enables processing million of points in a very efficient way.
In most cases, this method is especially cheap regarding computation time, since the distances have already been computed to find the K-nearest neighbors.
Compared to pure random sampling, it produces better uniformly distributed points at a coarser resolution and more likely preserves important areas.
In addition, it still includes randomness that is preferred in training of deep neural networks to better prevent against over-fitting.
Figure~\ref{fig:subsampling} demonstrates this approach.
Note, how the chair legs are rarely existing in a randomly sub-sampled version, while IDISS preserves the overall structure.
 \section{Implementation}
\label{sec:implementation}

To enable building complete DNNs with the presented flex-convolution model we have implemented two specific layers in TensorFlow: \textit{flex-convolution} and \textit{flex-max-pooling}.
Profiling shows that a direct highly hand-tuned implementation in CUDA leads to a run-time which is in the range of regular convolution layers (based on cuDNN) during inference.

\subsection{Neighborhood Processing}
\label{subsec:neighborhood}
Both new layers require a known neighborhood for each incoming point.
For a fixed set of points, this neighborhood is computed once upfront based on an efficient kD-tree implementation and kept fixed. For each point, the $k$ nearest neighbors are stored as indices into the point list.
The set of indices is represented as a tensor and handed over to each layer.

The \textit{flex-convolution} layer merely implements the convolution with continuous locations as described in Eq.~\eqref{eq:ours}.
Access to the neighbors follows the neighbor indices to lookup their specific feature vectors and location.
No data duplication is necessary.
As all points have the same number of neighbors, this step can be parallelized efficiently.
In order to make the position of each point available in each layer of the network, we attach the point location $\ell$ to each feature vector.

The \textit{flex-max-pooling} layer implements max-pooling over each point neighborhood individually, just like the grid-based version but without subsampling.

For subsampling, we exploit the IDISS approach described in Section~\ref{sec:sub-Sampling}.
Hereby, flex-max-pooling is applied before the subsampling procedure.
For the subsequent, subsampled layers the neighborhoods might have changed, as they only include the subsampled points.
As the point set is static and known beforehand, all neighborhood indices at each resolution can be computed on-the-fly during parallel data pre-fetching, which is  neglectable compared to the cost of a network forward+backward pass under optimal GPU utilization.

Upsampling (\textit{flex-upsampling}) is done by copying the features of the selected points into the larger-sized layer, initializing all other points with zero, like zero-padding in images and performing the flex-max-pooling operation.

\begin{table}[t]
  \caption{Profiling information of diverse implementations with 8 batch of $4096$ points with $C'=C=64$ and 9 neighbors using a CUDA profiler.}
  \label{tab:timings}
  \centering

  \begin{tabular}{l>{\raggedleft\arraybackslash}p{1.85cm}>{\raggedleft\arraybackslash}p{1.85cm}c>{\raggedleft\arraybackslash}p{1.85cm}>{\raggedleft\arraybackslash}p{1.85cm}}
  \toprule
                                         & \multicolumn{2}{c}{Timing} && \multicolumn{2}{c}{Memory}  \\
                                         \cline{2-3} \cline{5-6}
  Method                                           & Forward & Backward && Forward   & Backward   \\
  \midrule
  flex-convolution (pure TF)*                      & 1829ms  & 2738ms   && 34015.2MB & 63270.8MB \\
  flex-convolution (Ours)                       & 24ms    & 265ms    && 8.4MB     & 8.7MB \\
  flex-convolution (TC \cite{tensorcomprehension}) & 42ms    & -        && 8.4MB     & -\\
  \midrule
  grid-based conv.(cuDNN)           & 16ms    & 1.5ms    && 1574.1MB  & 153.4MB \\
  \midrule
  flex-max-pooling (Ours)                       & 1.44ms  & 15us     && 16.78MB   & 8.4MB \\
  \bottomrule
  \end{tabular}
\end{table}
\subsection{Efficient Implementation of Layer Primitives}
\label{subsec:implementation}
To ensure a reasonably fast training time, highly efficient GPU-implementations of flex-convolution and flex-max-pooling as a custom operation in TensorFlow are required.
We implemented a generic but hand-tuned CUDA operation, to ensure
optimal GPU-throughput.
Table~\ref{tab:timings} compares our optimized CUDA kernel against a version (pure TF) containing exclusively existing operations provided by the TensorFlow framework itself and its grid-based counterpart in cuDNN~\cite{cudnn} using the CUDA profiler for a \textit{single} flex-convolution layer on a set of parameters, which fits typical consumer hardware (Nvidia GTX 1080Ti).
As the grid-based convolution layer typically uses a kernel-size of $3\times 3\times C$ in the image domain, we set $k=9$ as well -- though we use $k=8$ in all subsequent point cloud experiments.
We did some experiments with a quite recent polyhedral compiler optimization using TensorComprehension (TC)~\cite{tensorcomprehension} to automatically tune a flex-convolution layer implementation.
While this approach seems promising, the lack of supporting flexible input sizes and slower performance currently prevents us from using these automatically generated CUDA kernels in practice.

An implementation of the flex-convolution layer by just relying on operations provided by the TensorFlow framework requires data duplication. We had to spread the pure TensorFlow version across 8 GPUs to run a \textit{single} flex-convolution layer.
Typical networks usually consist of several such operations.
Hence, it is inevitable to recourse on tuning custom implementations when applying such a technique to larger datasets.
Table~\ref{tab:timings} reveals that the grid-based version (cuDNN) prepares intermediate values in the forward pass resulting in larger memory consumption and faster back-propagation pass --- similar to our flex-max-pooling.

\subsection{Network Architecture for large-scale Semantic Segmentation}
\label{subsec:network}
With the new layers at hand, we can directly transfer the structure of existing image processing networks to the task of processing large point clouds. We will elaborate on our network design and choice of parameters for the task of semantic point cloud segmentation in more detail. Here, we draw inspiration from established hyper-parameter choices in 2D image processing.

\begin{figure}[tb]
  \centering
  \includegraphics[width=\textwidth]{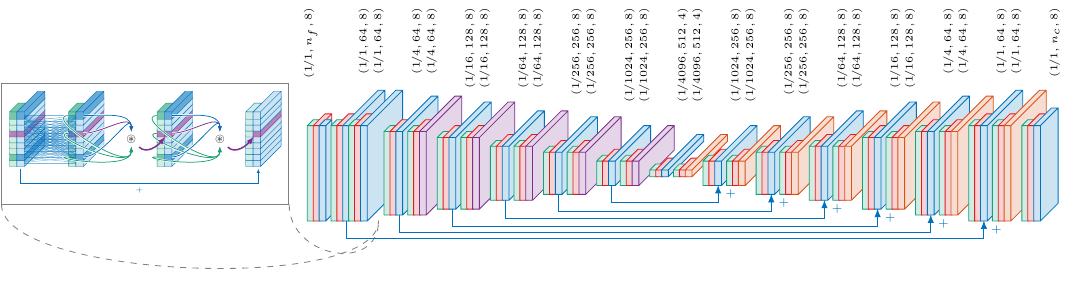}
  \caption{Network architecture for semantic 3D point cloud segmentation. The annotations $(a, d_f, k)$ represent the spatial resolution factor $a$ (\ie using $a\cdot n$ points) and feature length $d_f$ with $n_f$ input features and $n_c$ classes. The used neighborhood size is given by $k$. In each step, the position information \positionop and neighborhood information \neighbop is required besides the actual learned features.
  After flex-convolution layers \flexconvlayer, each downsampling step \flexmaxlayer (flex-max-pool) has a skip-connection to the corresponding decoder block with flex-upsampling layer \flexdeconvlayer.}
  \label{fig:network}
\end{figure}
Our network architecture follows the SegNet-Basic network~\cite{segnet} (a 2D counterpart for semantic image segmentation) with added U-net skip-connections~\cite{unet}.
It has a typical encoder-decoder network structure followed by a final point-wise soft-max classification layer.
To not obscure the effect of the flex-convolution layer behind several other effects, we explicitly do \textit{not} use tricks like Batch-Normalization, weighted soft-max classification, or computational expensive pre- resp.
post-processing approaches, which are known to enhance the prediction quality and could further be applied to the results presented in the Section~\ref{sec:experiments}.

The used architecture and output sizes are given in Figure~\ref{fig:network}.
The encoder network is divided into six stages of different spatial resolutions to process multi-scale information from the input point cloud.
Each resolution stage consists of two ResNet-blocks.
Such a ResNet block chains the following operations: $1\x 1$-convolution, flex-convolution, flex-convolution (compare Figure~\ref{fig:network}).
Herewith, the output of the last flex-convolution layer is added to the incoming feature following the common practice of Residual Networks \cite{resnet}.
To decrease the point cloud resolution across different stages, we add a flex-max-pooling operation with subsampling as the final layer in each stage of the encoder.
While a grid-based max-pooling is normally done with stride 2 in $x/y$ dimension, we use the flex-max-pooling layer to reduce the resolution $n$ by factor 4.
When the spatial resolution decreases, we increase the feature-length by factor two.

Moreover, we experimented with different neighborhood sizes $k$ for the flex-convolution layers.
Due to speed considerations and the widespread adoption of $3\times 3$ filter kernels in image processing we stick to a maximal nearest neighborhood size of $k=8$ in all flex-convolution layers.
We observed no decrease in accuracy against $k=16$ but a drop in speed by factor 2.2 for 2D-3D-S \cite{3d2dsdata}.

The decoder network mirrors the encoder architecture.
We add skip connections \cite{unet} from each stage in the encoder to its related layer in the decoder.
Increasing spatial resolution at the end of each stage is done via flex-upsampling.
We tested a trainable flex-transposed-convolution layer in some preliminary experiments and observed no significant improvements.
Since pooling irregular data is more light-weight (see Table~\ref{tab:timings}) regarding computation effort, we prefer this operation.
As this is the first network being able to process point clouds in such a large-scale setting, we expect choosing more appropriate hyper-parameters is possible when investing more computation time.
 \section{Experiments}
\label{sec:experiments}

We conducted several experiments to validate our approach.
These show that our flex-convolution-based neural network yields competitive performance to previous work on synthetic data for single object classification (\cite{modelnet40}, 1024 points) using fewer resources and provide some insights about human performance on this dataset. We improve single instance part segmentation (\cite{Yi16}, 2048 points).
Furthermore, we demonstrate the effectiveness of our approach by performing semantic point cloud segmentation on a large-scale real-world 3D scan (\cite{3d2dsdata}, 270~Mio. points) improving previous methods in both accuracy and speed.

\begin{wraptable}{r}{0.41\textwidth}
% \vspace{-1cm}
  \caption{Classification accuracy on ModelNet40 (1024 points) and 256 points${}^\ast$.}
  \label{tab:model40}
  \resizebox{\linewidth}{!}{
  \begin{tabular}{lcr}
  \toprule
  Method & Accuracy & \#params.\\
  \midrule
  PointNet \cite{pointnet} & 89.2 & 1'622'705\\
  PointNet2 \cite{pointnet2} & 90.7 & 1'658'120\\
  KD-Net\cite{kdnetwork} & 90.6 & 4'741'960\\
  D-FilterNet \cite{dyncedgefilters} & 87.4 & 345'288\\
  \midrule
  Human & 64.0& -\\
  \midrule
  Ours& 90.2 & 346'409\\
  Ours (1/4)&89.3 & 171'048\\
  \bottomrule
  \end{tabular}
   }
  % \vspace{-0.5cm}
\end{wraptable}

\subsection{Synthetic Data}
To evaluate the effectiveness of our approach, we participate in two benchmarks that arise from the ShapeNet~\cite{shapenet2015} dataset, which consists of synthetic 3D models created by digital artists.

\textbf{ModelNet40}~\cite{modelnet40} is a single object classification task of 40 categories.
We applied a smaller version of the previously described encoder network-part followed by a fully-connected layer and a classification layer.
Following the official test-split \cite{pointnet} of randomly sampled points from the object surfaces for object classification, we compare our results in Table \ref{tab:model40}.
Our predictions are provided from by a single forward-pass in contrast to a voting procedure as in the KD-Net \cite{kdnetwork}.
This demonstrates that a small flex-convolution neural network with significant fewer parameters provides competitive results on this benchmark set.
Even when using just 1/4th of the point cloud and thus an even smaller network the accuracy remains competitive.
To put these values in a context to human perception, we conducted a user study
asking participants to classify point clouds sampled from the official test split.
We allowed them to rotate the presented point cloud for the task of classification without a time limit.
Averaging all 2682 gathered object classification votes from humans reveals some difficulties with this dataset.
This might be related to the relatively unconventional choice of categories in the dataset, \ie plants and their flower pots and bowls are sometimes impossible to separate.
Please refer to the Supplementary for a screenshot of the user study, a confusion matrix, saliency maps and an illustration of label ambiguity.
\begin{table}[t]
\caption{ShapeNet part segmentation results per category and mIoU (\%) for different methods and inference speed (on a Nvidia GeForce GTX 1080 Ti). }
\label{tab:shapenetIoU}
\centering
\definecolor{Gray}{gray}{0.90}
\definecolor{LightCyan}{rgb}{0.88,1,1}
\newcolumntype{a}{>{\columncolor{Gray}}c}
\resizebox{\linewidth}{!}{
\begin{tabular}{l|cccccccccccccccc|aa}
\toprule
                               & Airpl.        & Bag           & Cap           & Car           & Chair         & Earph.        & Guitar        & Knife         & Lamp          & Laptop        & Motorb.       & Mug           & Pistol        & Rocket        & Skateb.       & Table         & {mIoU}        & shapes/sec   \\
\hline
Kd-Network \cite{kdnetwork}    & 80.1          & 74.6          & 74.3          & 70.3          & 88.6          & 73.5          & 90.2          & \textbf{87.2} & 81.0          & 94.9          & 57.4          & 86.7          & 78.1          & 51.8          & 69.9          & 80.3          & 77.4          & n.a.\\
PointNet \cite{pointnet}       & 83.4          & 78.7          & 82.5          & 74.9          & 89.6          & 73.0          & 91.5          & 85.9          & 80.8          & 95.3          & 65.2          & 93.0          & 81.2          & 57.9          & 72.8          & 80.6          & 80.4          & n.a. \\
PointNet++ \cite{pointnet2}    & 82.4          & 79.0          & 87.7          & 77.3          & \textbf{90.8} & 71.8          & 91.0          & 85.9          & 83.7          & 95.3          & 71.6          & 94.1          & 81.3          & 58.7          & 76.4          & 82.6          & 81.9          & 2.7 \\
SPLATNet3D \cite{su18splatnet} & 81.9          & 83.9          & 88.6          & \textbf{79.5} & 90.1          & 73.5          & 91.3          & 84.7          & \textbf{84.5} & 96.3          & 69.7          & 95.0          & 81.7          & 59.2          & 70.4          & 81.3          & 82.0          & 9.4   \\
SGPN \cite{wang2018sgpn}       & 80.4          & 78.6          & 78.8          & 71.5          & 88.6          & \textbf{78.0} & 90.9          & 83.0          & 78.8          & 95.8          & \textbf{77.8} & 93.8          & \textbf{87.4} & 60.1          & \textbf{92.3} & \textbf{89.4} & 82.8          & n.a. \\
\hline
Ours                           & \textbf{83.6} & \textbf{91.2} & \textbf{96.7} & \textbf{79.5} & 84.7          & 71.7          & \textbf{92.0} & 86.5          & 83.2          & \textbf{96.6} & 71.7          & \textbf{95.7} & 86.1          & \textbf{74.8} & 81.4          & 84.5          & \textbf{85.0} & \textbf{489.3} \\
\bottomrule
\end{tabular}
}
\end{table}

\begin{figure}[tb]
  \centering
  \includegraphics[]{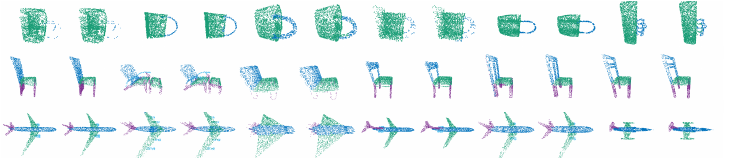}
  \caption{Our semantic segmentation results on ShapeNet (ground-truth (left), prediction (right)) pairs.
  Please refer to the supplementary for more results at higher resolution.}
  \label{fig:shapenet_results}
\end{figure}

\textbf{ShapeNet Part Segmentation}~\cite{Yi16} is a semantic segmentation task with per-point annotations of 31963 models separated into 16 shape categories.
We applied a smaller version of the previously described segmentation network that receives the $(x,y,z)$ position of $2048$ points per object.
For the evaluation, we follow the procedure of~\cite{su18splatnet} by training a network for per category.
Table~\ref{tab:shapenetIoU} contains a comparison of methods using only point cloud data as input.
Our method demonstrates an improvement of the average mIoU while being able to process a magnitude more shapes per second. Examples of ShapeNet part segmentation are illustrated in Figure~\ref{fig:shapenet_results}.
These experiments on rather small synthetic data confirm our hypothesis that even in three dimensions simple filters with a small amount of learnable parameters are sufficient in combination with deeper network architectures.
This matches with the findings that are known from typical CNN architectures of preferring deeper networks with small \(3\times3\) filters.
The resulting smaller memory footprint and faster computation time enable processing more points in reasonable time.
We agree with \cite{su18splatnet,wang2018sgpn} on the data labeling issues.

\subsection{Real-World Semantic Point Cloud Segmentation}
\label{subsec:segmentation}

% \vspace{-0.2cm}
To challenge our methods at scale, we applied the described network from Section~\ref{subsec:network} to the 2D-3D-S dataset \cite{3d2dsdata}.
This real-world dataset covers 3D scanning information from six square kilometers of several building complexes collected by a Matterport Camera.
Previous approaches are based on sliding windows, either utilizing hand-crafted feature, \eg local curvature, occupancy and point density information per voxel \cite{3d2dsresults,3d2dsdata} or process small sub-sampled chunks PointNet \cite{pointnet}, SGPN~\cite{wang2018sgpn} (4096 points, Figure~\ref{fig:teaser}).
We argue, that a neural network as described in Section~\ref{subsec:network} can learn all necessary features directly from the data -- just like in the 2D case and at \textit{full} resolution.

An ablation study on a typical room reveals the effect of different input features $f$.
Besides neighborhood information, providing only constant initial features $f=1$ yields $0.31$ mAP.
Hence, this is already enough information to perform successful semantic segmentation.
To account for the irregularity in the data, it is however useful to use normalized position data $f=(1,x,y,z)$ besides the color information $f=(1,x,y,z,r,g,b)$ which increases the accuracy to $0.39$ mAP resp. $0.50$ mAP.
Our raw network predictions from a single inference forward pass out-performs previous approaches given the same available information and approaches using additional input information but lacks precision in categories like beam, column, and door, see Table~\ref{tab:stanford_AP}.
% While improving state-of-the-art results on 2D-3D-S our network lacks precision in categories like beam, column, and door.
Providing features like local curvature besides post-processing \cite{3d2dsdata} greatly simplify detecting these kinds of objects.
Note, our processing of point clouds at full resolution benefits the handling of smaller objects like chair, sofa and table.

\begin{figure}[tb]
  \centering
  \includegraphics[width=\textwidth]{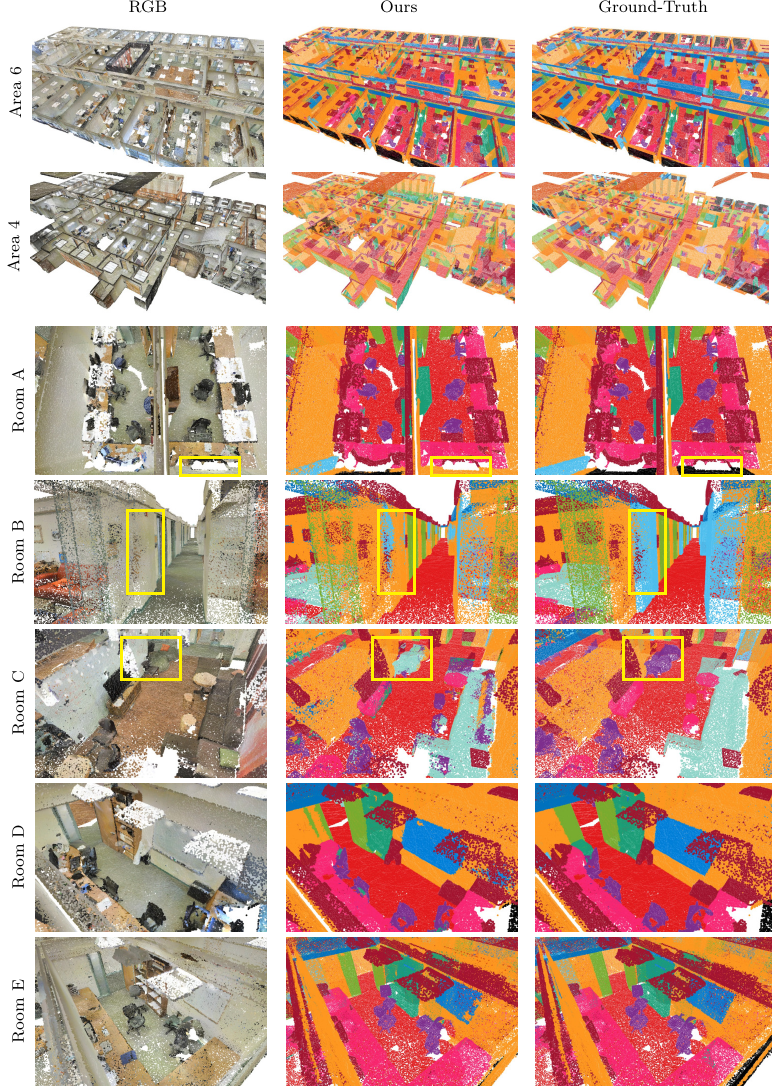}
  \caption{Semantic point cloud segmentation produced as raw outputs of our proposed network from the held-out validation set.
  In this point-based rendering, surfaces might not be illustrated as opaque.}
  \label{fig:final-results}
\end{figure}

Consider Figure~\ref{fig:final-results}, the highlighted window region in room A is classified as wall because the blinds are closed, thus having a similar appearance.
In room B, our network miss-classifies the highlighted column as ``wall'', which is not surprising as both share similar geometry and color.
Interestingly, in room C our network classifies the beanbag as ``sofa'', while its ground-truth annotation is ``chair''.
For more results please refer to the accompanying video.

Training is done on two Nvidia GTX 1080Ti with batch-size $16$ for two days on point cloud chunk with $128^2$ points using the Adam-Optimizer with learning-rate $3\cdot 10^{-3}$.
\clearpage

To benchmark inference, we compared ours against the author's implementations of previous work \cite{pointnet,pointnet2,wang2018sgpn} on different point clouds sizes $n$.
Memory requirements limits the number of processed points to at most $131$k \cite{pointnet2}, $500$k \cite{wang2018sgpn}, $1$Mio \cite{pointnet} points (highlighted region in Figure~\ref{fig:teaser}).
We failed to get meaningful performance in terms of accuracy from these approaches when increasing $n>4096$.
In contrast, ours -- based on a fully convolutional network -- can process up to $7$ Mio. points concurrently providing the \textit{same} performance during inference within 4.7 seconds. Note, \cite{pointnet} can at most process 1 Mio. points within 7.1 seconds.
Figure~\ref{fig:teaser} further reveals an exponential increase of runtime for the PointNet family \cite{pointnet,pointnet2,wang2018sgpn}, ours provides significant faster inference and shows better utilization for larger point clouds with a linear increase of runtime.

\begin{table}[tb]
  \caption{Class specific average precision (AP) on the 2D-3D-S dataset.
  (\ddag) uses additional input features like local curvature, point densities, surface normals.
  (*) uses non-trivial post-processing and (**) a mean filter post-processing.}
  \label{tab:stanford_AP}
  \centering

  \definecolor{Gray}{gray}{0.90}
  \definecolor{LightCyan}{rgb}{0.88,1,1}
  \newcolumntype{a}{>{\columncolor{Gray}}c}
  \resizebox{\linewidth}{!}{
  \begin{tabular}{l|ccccc|ccccccc|a}
      \toprule
                                         & Table          & Chair          & Sofa           & Bookc.         & Board          & Ceiling        & Floor          & Wall           & Beam           & Col.           & Wind.          & Door           & {mAP}  \\
                              \hline
     Armenin \etal \cite{3d2dsdata}*     & 46.02          & 16.15          & 6.78           & {54.71}        & 3.91           & 71.61          & 88.70          & 72.86          & 66.67          & \textbf{91.77} & 25.92          & 54.11          & {{49.93}}\\
     Armenin \etal \cite{3d2dsdata}\ddag & 39.87          & 11.43          & 4.91           & \textbf{57.76} & 3.73           & 50.74          & 80.48          & 65.59          & {68.53}        & 85.08          & 21.17          & 45.39          & {44.19}\\
     PointNet \cite{pointnet}*           & 46.67          & 33.80          & 4.76           & n.a.           & 11.72          & n.a.           & n.a.           & n.a.           & n.a.           & n.a.           & n.a.           & n.a.           & n.a.    \\
     SGPN \cite{wang2018sgpn}*           & 46.90          & 40.77          & 6.38           & 47.61          & 11.05          & 79.44          & 66.29          & \textbf{88.77} & \textbf{77.98} & 60.71          & \textbf{66.62} & \textbf{56.75} & {{54.35}} \\
     \hline
     Ours                                & {66.03}        & {51.75}        & {15.59}        & 39.03          & {43.50}        & {87.20}        & {96.00}        & 65.53          & 54.76          & 52.74          & {55.34}        & 35.81          & {{55.27}}\\
     Ours**                              & \textbf{67.02} & \textbf{52.75} & \textbf{16.61} & {39.26}        & \textbf{47.68} & \textbf{87.33} & \textbf{96.10} & {65.52}        & {56.83}        & {55.10}        & {57.66}        & {36.76}        & {\textbf{56.55}}\\
      \bottomrule
      \end{tabular}
  }
      \vspace*{-0.5cm}

\end{table}
 % \vspace{-0.1cm}
\textbf{Limitation}
As we focus on static point cloud scans ours is subject to the same limitations as \cite{pointnet,pointnet2,su18splatnet,wang2018sgpn}, where neighborhoods are computed during parallel data pre-fetching.
Handling dynamic point clouds, \eg completion or generation, requires an approximate nearest-neighborhood layer.
Our prototype implementation suggests this can be done \textit{within} the network.
For 2 Million points it takes around 1 second which is still faster by a factor of 8 compared to the used kd-Tree, which has neglectable costs being part of parallel pre-fetching.
% \vspace{-0.5cm}
\section{Conclusion}
We introduced a novel and natural extension to the traditional convolution, transposed convolution and max-pooling primitives for processing irregular point sets.
The novel sparse operations work on the local neighborhood of each point, which is provided by indices to the $k$ nearest neighbors. Compared to 3D CNNs our approach can be extended to support even high-dimensional point sets easily.
As the introduced layers behave very similar to convolution layers in networks designed for 2D image processing, we can leverage the full potential of already successful architectures. This is against recent trends in point cloud processing with highly specialized architectures which sometimes rely on hand-crafted input features, or heavy pre- and post-processing.
We demonstrate state-of-the-art results on small synthetic data as well as large real-world datasets while processing millions of points concurrently and efficiently.
% At the same time, our approach can concurrently process millions of points while promising a fast inference.
\begin{flushright}
{\scriptsize\noindent This work was supported by the DFG: SFB 1233, Robust Vision.}
\end{flushright}

\clearpage

\bibliographystyle{splncs04}
\bibliography{egbib}

\begin{thebibliography}{10}
\providecommand{\url}[1]{\texttt{#1}}
\providecommand{\urlprefix}{URL }
\providecommand{\doi}[1]{https://doi.org/#1}

\bibitem{3d2dsresults}
{Armeni}, I., {Sax}, A., {Zamir}, A.R., {Savarese}, S.: {Joint 2D-3D-Semantic
  Data for Indoor Scene Understanding}. ArXiv e-prints  (Feb 2017)

\bibitem{3d2dsdata}
Armeni, I., Sener, O., Zamir, A.R., Jiang, H., Brilakis, I., Fischer, M.,
  Savarese, S.: 3d semantic parsing of large-scale indoor spaces. In:
  Proceedings of the IEEE Conference on Computer Vision and Pattern Recognition
  (CVPR) (2016)

\bibitem{segnet}
Badrinarayanan, V., Kendall, A., Cipolla, R.: Segnet: A deep convolutional
  encoder-decoder architecture for image segmentation. IEEE Transactions on
  Pattern Analysis and Machine Intelligence (PAMI)  (2017)

\bibitem{sphericalprojection}
Cao, Z., Huang, Q., Karthik, R.: 3d object classification via spherical
  projections. In: International Conference on 3D Vision (3DV). pp. 566--574.
  IEEE (2017)

\bibitem{shapenet2015}
Chang, A.X., Funkhouser, T., Guibas, L., Hanrahan, P., Huang, Q., Li, Z.,
  Savarese, S., Savva, M., Song, S., Su, H., Xiao, J., Yi, L., Yu, F.:
  {ShapeNet: An Information-Rich 3D Model Repository}. Tech. Rep.
  arXiv:1512.03012 [cs.GR], Stanford University --- Princeton University ---
  Toyota Technological Institute at Chicago (2015)

\bibitem{cudnn}
Chetlur, S., Woolley, C., Vandermersch, P., Cohen, J., Tran, J., Catanzaro, B.,
  Shelhamer, E.: cudnn: Efficient primitives for deep learning. CoRR  (2014)

\bibitem{dynamicfilternetwork}
De~Brabandere, B., Jia, X., Tuytelaars, T., Van~Gool, L.: Dynamic filter
  networks. In: Advances in Neural Information Processing Systems (NIPS) (2016)

\bibitem{DBLP:conf/dagm/GrohRL17}
Groh, F., Resch, B., Lensch, H.P.A.: Multi-view continuous structured light
  scanning. In: Pattern Recognition - 39th German Conference, {GCPR} 2017,
  Basel, Switzerland, September 12-15, 2017, Proceedings. pp. 377--388 (2017).
  \doi{10.1007/978-3-319-66709-6\_30}

\bibitem{resnet}
He, K., Zhang, X., Ren, S., Sun, J.: Identity mappings in deep residual
  networks. In: Proceedings of the European Conference on Computer Vision
  (ECCV). pp. 630--645 (2016)

\bibitem{hermosilla2018monte}
Hermosilla, P., Ritschel, T., V{\'a}zquez, P.P., Vinacua, {\`A}., Ropinski, T.:
  Monte carlo convolution for learning on non-uniformly sampled point clouds.
  arXiv preprint arXiv:1806.01759  (2018)

\bibitem{audioprocessing}
Hershey, S., Chaudhuri, S., Ellis, D.P.W., Gemmeke, J.F., Jansen, A., Moore,
  C., Plakal, M., Platt, D., Saurous, R.A., Seybold, B., Slaney, M., Weiss, R.,
  Wilson, K.: Cnn architectures for large-scale audio classification. In: IEEE
  International Conference on Acoustics, Speech and Signal Processing (ICASSP)
  (2017)

\bibitem{spatialtransformer}
Jaderberg, M., Simonyan, K., Zisserman, A., kavukcuoglu, k.: Spatial
  transformer networks. In: Cortes, C., Lawrence, N.D., Lee, D.D., Sugiyama,
  M., Garnett, R. (eds.) Advances in Neural Information Processing Systems
  (NIPS), pp. 2017--2025. Curran Associates, Inc. (2015)

\bibitem{kipf2017semi}
Kipf, T.N., Welling, M.: Semi-supervised classification with graph
  convolutional networks. In: International Conference on Learning
  Representations (ICLR) (2017)

\bibitem{kdnetwork}
Klokov, R., Lempitsky, V.: Escape from cells: Deep kd-networks for the
  recognition of 3d point cloud models. In: Proceedings of the IEEE
  International Conference on Computer Vision (ICCV). pp. 863--872 (10 2017)

\bibitem{winograd}
Lavin, A., Gray, S.: Fast algorithms for convolutional neural networks. In:
  Proceedings of the IEEE Conference on Computer Vision and Pattern Recognition
  (CVPR). pp. 4013--4021 (2016)

\bibitem{voxnet}
Maturana, D., Scherer, S.: {VoxNet: A 3D Convolutional Neural Network for
  Real-Time Object Recognition}. In: International Conference on Intelligent
  Robots and Systems (2015)

\bibitem{pointnet}
Qi, C.R., Su, H., Mo, K., Guibas, L.J.: Pointnet: Deep learning on point sets
  for 3d classification and segmentation. Proceedings of the IEEE Conference on
  Computer Vision and Pattern Recognition (CVPR)  (2017)

\bibitem{qi2016volumetric}
Qi, C.R., Su, H., Niessner, M., Dai, A., Yan, M., Guibas, L.J.: Volumetric and
  multi-view cnns for object classification on 3d data. Proceedings of the IEEE
  Conference on Computer Vision and Pattern Recognition (CVPR)  (2016)

\bibitem{pointnet2}
Qi, C.R., Yi, L., Su, H., Guibas, L.J.: Pointnet++: Deep hierarchical feature
  learning on point sets in a metric space. In: Guyon, I., Luxburg, U.V.,
  Bengio, S., Wallach, H., Fergus, R., Vishwanathan, S., Garnett, R. (eds.)
  Advances in Neural Information Processing Systems (NIPS), pp. 5099--5108.
  Curran Associates, Inc. (2017)

\bibitem{octnet}
Riegler, G., Ulusoy, A.O., Bischof, H., Geiger, A.: Octnetfusion: Learning
  depth fusion from data. In: International Conference on 3D Vision (3DV) (Oct
  2017)

\bibitem{unet}
Ronneberger, O., P.Fischer, Brox, T.: U-net: Convolutional networks for
  biomedical image segmentation. In: Medical Image Computing and
  Computer-Assisted Intervention (MICCAI). LNCS, vol.~9351, pp. 234--241.
  Springer (2015)

\bibitem{PANORAMA}
Sfikas, K., Pratikakis, I., Theoharis, T.: Ensemble of panorama-based
  convolutional neural networks for 3d model classification and retrieval.
  Computers and Graphics  (2017)

\bibitem{dyncedgefilters}
Simonovsky, M., Komodakis, N.: Dynamic edge-conditioned filters in
  convolutional neural networks on graphs. In: Proceedings of the IEEE
  Conference on Computer Vision and Pattern Recognition (CVPR) (2017),
  \url{https://arxiv.org/abs/1704.02901}

\bibitem{vggnet}
Simonyan, K., Zisserman, A.: Very deep convolutional networks for large-scale
  image recognition. CoRR  (2014)

\bibitem{su18splatnet}
Su, H., Jampani, V., Sun, D., Maji, S., Kalogerakis, E., Yang, M.H., Kautz, J.:
  {SPLATN}et: Sparse lattice networks for point cloud processing. In:
  Proceedings of the IEEE Conference on Computer Vision and Pattern Recognition
  (CVPR). pp. 2530--2539 (2018)

\bibitem{SuMKL15}
Su, H., Maji, S., Kalogerakis, E., Learned{-}Miller, E.G.: Multi-view
  convolutional neural networks for 3d shape recognition. In: Proceedings of
  the IEEE International Conference on Computer Vision (ICCV) (2015)

\bibitem{tensorcomprehension}
Vasilache, N., Zinenko, O., Theodoridis, T., Goyal, P., DeVito, Z., Moses,
  W.S., Verdoolaege, S., Adams, A., Cohen, A.: Tensor comprehensions:
  Framework-agnostic high-performance machine learning abstractions (2018)

\bibitem{wang2018sgpn}
Wang, W., Yu, R., Huang, Q., Neumann, U.: Sgpn: Similarity group proposal
  network for 3d point cloud instance segmentation. In: Proceedings of the IEEE
  Conference on Computer Vision and Pattern Recognition (CVPR). pp. 2569--2578
  (2018)

\bibitem{iccv2017/Wieschollek}
Wieschollek, P., Sch{\"{o}}lkopf, M.H.B., Lensch, H.P.A.: Learning blind motion
  deblurring. In: International Conference on Computer Vision (ICCV) (October
  2017)

\bibitem{modelnet40}
Wu, Z., Song, S., Khosla, A., Yu, F., Zhang, L., Tang, X., Xiao, J.: 3d
  shapenets: A deep representation for volumetric shapes. In: Proceedings of
  the IEEE Conference on Computer Vision and Pattern Recognition (CVPR). pp.
  1912--1920 (2015)

\bibitem{Zhirong15CVPR}
Wu, Z., Song, S., Khosla, A., Yu, F., Zhang, L., Tang, X., Xiao, J.: 3d
  shapenets: A deep representation for volumetric shapes. In: Proceedings of
  the IEEE Conference on Computer Vision and Pattern Recognition (CVPR) (2015)

\bibitem{Yi16}
Yi, L., Kim, V.G., Ceylan, D., Shen, I.C., Yan, M., Su, H., Lu, C., Huang, Q.,
  Sheffer, A., Guibas, L.: A scalable active framework for region annotation in
  3d shape collections. ACM Transactions on Graphics (SIGGRAPH ASIA)  (2016)

\end{thebibliography}

\end{document}